\documentclass{article}
\usepackage{spconf,amsmath,graphicx}

\newcommand{\figref}[1]{Figure~\ref{#1}}
\newcommand{\tabref}[1]{Table~\ref{#1}}

\usepackage{booktabs}
\usepackage{threeparttable}
\usepackage{multirow}
\usepackage{color}

\usepackage{epsfig}
\usepackage{graphicx}
\usepackage{subfigure}
\usepackage{overpic}
\usepackage{amsfonts}
\usepackage{bm}
\usepackage{amsmath}
\usepackage{bbm}
\usepackage{url}
\usepackage{cleveref}
\crefformat{section}{\S#2#1#3}
\crefformat{subsection}{\S#2#1#3}

\title{Improving Cross-Modal Understanding in Visual Dialog \\via Contrastive Learning}
\name{Feilong Chen$^{1,2}$, Xiuyi Chen$^{1,3}$, Shuang Xu$^{1}$, Bo Xu$^{1,2,3}$}
\address{
$^{1}$Institute of Automation, Chinese Academy of Sciences, Beijing, China\\
$^{2}$School of Future Technology, $^{3}$University of Chinese Academy of Sciences, Beijing, China\\
\small \tt \{chenfeilong2018, chenxiuyi2017, shuang.xu, xubo\}@ia.ac.cn}

\begin{document}
\ninept
\maketitle
\begin{abstract}
Visual Dialog is a challenging vision-language task since the visual dialog agent needs to answer a series of questions after reasoning over both the image content and dialog history. Though existing methods try to deal with the cross-modal understanding in visual dialog, they are still not enough in ranking candidate answers based on their understanding of visual and textual contexts. In this paper, we analyze the cross-modal understanding in visual dialog based on the vision-language pre-training model VD-BERT and propose a novel approach to improve the cross-modal understanding for visual dialog, named ICMU. ICMU enhances cross-modal understanding by distinguishing different pulled inputs (i.e. pulled images, questions or answers) based on four-way contrastive learning.  In addition, ICMU exploits the single-turn visual question answering to enhance the visual dialog model's cross-modal understanding to handle a multi-turn visually-grounded conversation. Experiments show that the proposed approach improves the visual dialog model's cross-modal understanding and brings satisfactory gain to the VisDial dataset.
\end{abstract}
\section{Introduction}
\label{sec:intro}

\begin{figure}[t]
\centering
\scalebox{0.95}{
  \begin{overpic}[width=\columnwidth]{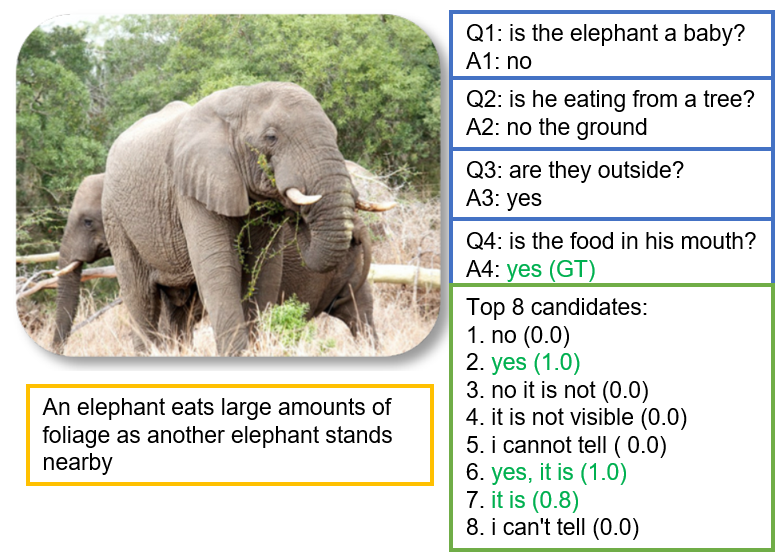}
  \end{overpic}
  }
  \caption{A motivating example of cross-modal understanding of VD-BERT~\cite{wang2020vd}. We show the candidates ranking results of VD-VBERT based on its cross-modal understanding. It can be seen that in the first 8 candidates, wrong answers account for most of them, and the ranking results of correct answers are not so good.
  }\label{fig:example} \vspace{-0.3cm}
\end{figure}

Recently, with the rise of pre-trained models~\cite{jiao2019tinybert}, researchers have begun to explore vision-and-language task~\cite{ren2015exploring,xu2015show,das2017visual} with pre-trained models~\cite{wang2020vd}. Specifically, visual dialog~\cite{Chen2021GoGRG,agarwal2020history,Chen2021MultimodalIT,Qi2020TwoCP}, which aims to hold a meaningful conversation with a human about a given image, is a challenging task that requires models have sufficient cross-modal understanding based on both visual and textual context to answer the current question.

One way to gain sufficient cross-modal understanding is through utilizing kinds of attention mechanism~\cite{lu2017best,wu2018you,Kottur2018VisualCR}. ReDAN~\cite{gan2019multi} and DMAM~\cite{chen2020dmrm} use multi-step reasoning based on dual attention to learn cross-modal understanding. DAN~\cite{guo2019dual}, MCAN~\cite{agarwal2020history} and LTMI~\cite{nguyenefficient} utilize multi-head attention mechanisms to manage multi-modal intersection. Moreover, there are some approaches~\cite{zheng2019reasoning,schwartz2019factor,jiang2020dualvd,guo2020iterative,jiang2020kbgn} using graph-based structures to learn cross-modal understanding. 

However, the approaches mentioned above do not utilize pre-trained models, which have a strong power to deal with vision-and-language tasks. Visdial-BERT~\cite{murahari2019large} and VD-BERT~\cite{wang2020vd} take advantage of the pre-trained model to greatly improve the performance of the visual dialog task. As shown in \figref{fig:example}, the SOTA model VD-BERT often makes mistakes and usually ranks the wrong answers first. VD-BERT does not have enough cross-modal understanding capabilities, so that it often scores unrelated wrong answers very high, such as the top 1 candidate answer ``\texttt{no}'' to the question Q4 ``\texttt{is the food in his mouth ?}'' shown in \figref{fig:example}.

\begin{figure*}[t]
\centering
\scalebox{0.90}{
  \begin{overpic}[width=\textwidth]{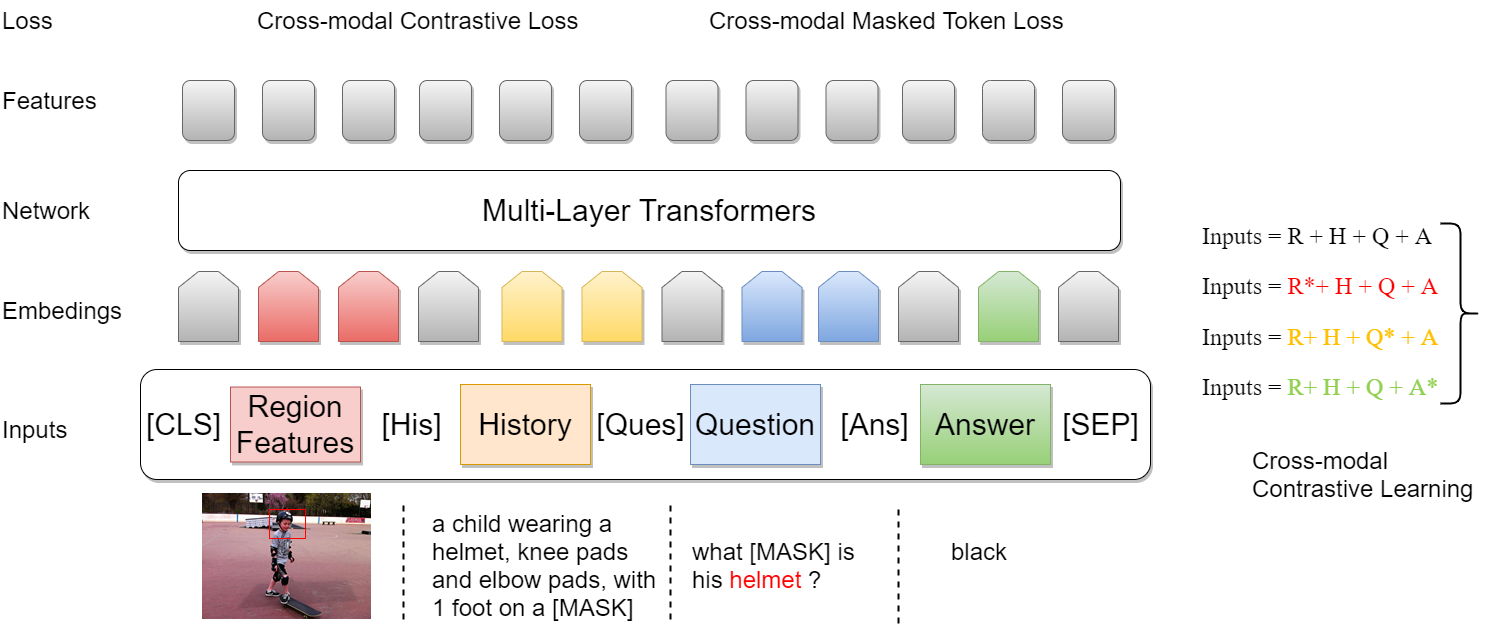}
  \end{overpic}
  }
  \caption{The Framework of our ICMU. * indicates the pulled inputs.
  }\label{fig:framework} 
\end{figure*}

In this paper, we propose a novel approach to improve the cross-modal understanding for visual dialog, named ICMU. ICMU enhances cross-modal understanding by distinguishing different pulled inputs (i.e. pulled images, questions or answers) based on four-way contrastive learning. What's more, ICMU exploits the single-turn visual question answering to enhance the visual dialog model's cross-modal understanding to handle a multi-turn visually-grounded conversation. Experiments show that the proposed approach improves the visual dialog model's cross-modal understanding and brings satisfactory gain on the VisDial dataset~\cite{das2017visual}. The contributions of this work are summarized as follows:

\begin{itemize}
  \item We propose a novel approach ICMU, including 4-way contrastive learning and enhancing by utilizing VQA, to improve the cross-modal understanding based on vision-and-language pre-trained models for visual dialog.
  \item We conduct extensive experiments and ablation studies on the large-scale datasets VisDial v1.0. Experimental results show that our approach improves the visual dialog model's cross-modal understanding and brings satisfactory gain.
\end{itemize}

\section{Methodology}

In this section, we first formally describe the visual dialog task. Given a current question $Q_t$ with an image $I$ at $t$-th turn, as well as its dialog history $H_t=\{C, (Q_1,A_1),...,(Q_{t-1},A_{t-1})\}$ (where $C$ denotes the image caption), the dialog model is required to predict its answer $A_t$ by ranking a list of $100$ answer candidates $\{\hat{A}_t^1, \hat{A}_t^2,...,\hat{A}_t^{100}\}$.

Figure~\ref{fig:framework} shows the overview of our approach. First, we employ a unified vision-dialog Transformer to encode both the image and dialog history, where we append an answer candidate $\hat{A}_t$ in the input to model their interactions in an early fusion manner. Next, we adopt cross-modal masked token loss and cross-modal contrastive loss to train the model for effective cross-modal understanding in visual dialog. In addition, we exploit the single-turn visual question answering to enhance the visual dialog model's cross-modal understanding to handle a multi-turn visually-grounded conversation.

\vspace{-0.4em}
\subsection{Vision-Dialog Transformer}\label{ssec:enc_pt}

\subsubsection{Visual Features.} 
Given an image $I$, we employ Faster R-CNN~\cite{DBLP:conf/nips/RenHGS15} pre-trained on Visual Genome~\cite{DBLP:journals/ijcv/KrishnaZGJHKCKL17} to extract the object-level vision features $R_I = \{o_1,...,o_k\}$, where each object feature $o_i$ is a $2048$-d Region-of-Interest (RoI) feature. $k$ is fixed to $36$ in our setting. In addition, we adopt normalized bounding box coordinates as the spatial location due to disorder of visual objects. Specifically, we define the location information by constructing a $5$-d vector: $p_i = (\frac{x_1}{W}, \frac{y_1}{H},\frac{x_2}{W}, \frac{y_2}{H}, \frac{(x_2-x_1)(y_2-y_1)}{WH})$, where $(x_1, y_1)$ and $(x_2, y_2)$ are the coordinates of the bottom-left and top-right corner of the $i$-th object, $W$ and $H$ respectively denote the width and height of the input image, and the last element is the relative area of the object. We also extend $p_i$ with its class id and confidence score for a richer representation to $7$-d vector.

\subsubsection{Textual Features.} 
For the textual features, we pack all the textual elements (the history, question and answer candidate) into a long sequence and employ WordPiece tokenizer~\cite{wu2016google} to split it into a word sequence $\mathbf{w}$, where each word is embedded with an absolute positional code following~\cite{DBLP:conf/naacl/DevlinCLT19}. 

\subsubsection{Cross-Modality Encoding.} 
Like a most vision-and-language transformers, we integrate the image objects with language elements into a whole input sequence.
As shown in~\figref{fig:framework}, we use some special tokens to segment different elements in the input sequence. We use \texttt{[CLS]} to denote the beginning of the sequence, and \texttt{[SEP]} to separate the two modalities. 
Moreover, we utilize a special token  \texttt{[HIS]} to denote \textit{end of turn}~\cite{DBLP:journals/corr/abs-1908-04812}, which informs the model when the dialog turn ends. And we use \texttt{[Ques]} and \texttt{[Ans]} to segment the current question and the answer candidate.
As such, we prepare the input sequence into the format
as $\mathbf{x}$ $=$ (\texttt{[CLS]}, $o_1,...,o_k$, \texttt{[SEP]}, $C$, \texttt{[His]}, $Q_1A_1$, \texttt{[His]}, ..., \texttt{[Ques]}, $Q_t$, \texttt{[Ans]}, $\hat{A}_t$, \texttt{[SEP]}). 
Finally, We combine each input token embedding with its position embedding and segment embedding ($0$ or $1$, indicating whether it is image or text) and then perform layer normalization~\cite{DBLP:journals/corr/BaKH16}.

\subsubsection{Transformer Backbone.}
We utilize transformer encoder as the Transformer backbone to handle cross-modal understanding. Formally, we denote the embedded vision-language inputs as $\mathbf{H}^0=[\mathbf{e}_1, ..., \mathbf{e}_{|\mathbf{x}|}]$ and then encode them into multiple levels of cross-modal representations $\mathbf{H}^l=[\mathbf{h}_1^l, ...,\mathbf{h}^l_{|\mathbf{x}|}]$ using $L$-stacked Transformer blocks, where the $l$-th Transformer block is denoted as $\mathbf{H}^l=\text{Transformer}(\mathbf{H}^{l-1}), l\in[1, L]$. Specifically, the cross-modal representations $\mathbf{H}^l$ is calculated by using the multi-head self-attention~\cite{DBLP:conf/nips/VaswaniSPUJGKP17} as follows:  
% the previous layer's output $\mathbf{H}^{l-1}\in\mathbb{R}^{|\mathbf{x}|\times d_h}$ is aggregated :
\begin{equation}
    \mathbf{Q}=\mathbf{H}^{l-1}\mathbf{W}_l^{Q}, \mathbf{K}=\mathbf{H}^{l-1}\mathbf{W}_l^{K}, \mathbf{V}=\mathbf{H}^{l-1}\mathbf{W}_l^{V},
    \vspace{-0.2in}
\end{equation}

\begin{equation}\label{eq:mask}
    \mathbf{M}_{ij}= \begin{cases}
      0, & \text{allow to attend},\\
      -\infty, & \text{prevent from attending},\\
    \end{cases}  
     \vspace{-0.1in}
\end{equation}

\begin{equation}
    \mathbf{A}_l=\text{softmax}(\frac{\mathbf{Q}\mathbf{K}^T}{\sqrt{d_k}}+\mathbf{M})\mathbf{V},
\end{equation}
where $\mathbf{W}_l^{Q}, \mathbf{W}_l^{K}, \mathbf{W}_l^{V}\in\mathbb{R}^{d_h\times d_k}$ are learnable weights for computing the queries, keys, and values respectively, and $\mathbf{M}\in\mathbb{R}^{|\mathbf{x}|\times|\mathbf{x}|}$ is the self-attention mask that determines whether tokens from two sources can attend each other.
Then $\mathbf{A}_l$ is passed into a feedforward layer to compute $\mathbf{H}^l$ for the next layer:
\begin{equation}\label{eq:mask}
    \mathbf{H}^l = {\rm FFN}(\mathbf{A}_l)
     \vspace{-0.1in}
\end{equation}

\begin{table}[t!]
    \centering
    \resizebox{0.98\columnwidth}!{
    \begin{tabular}{l|cccccc}
    \toprule
     Model & NDCG & MRR & R@1 & R@5 & R@10 & Mean \\\midrule
    ReDAN
    & 57.63 & 64.75 & 51.10  & 81.73  & 90.90  &  3.89  \\
    GNN-EM
    & 52.82 & 61.37 & 47.33 & 77.98 & 87.83 & 4.57 \\
    DualVD
    & 56.32 & 63.23 & 49.25 & 80.23 & 89.70 & 4.11 \\
    FGA
    & 56.90 & 66.20 & 52.75 & 82.92 & 91.07 & 3.80 \\
    CAG
    & 56.64 & 63.49 & 49.85 & 80.63 & 90.15 & 4.11 \\
    KBGN
    & 57.60 &  64.13 &  50.47 & 80.70 &  90.16 &  4.08 \\
    LG & 58.55 & 64.00 & 50.63 & 80.58 & 90.20 & 4.12 \\
    GoG & 60.38 & 63.13 & 49.88 & 79.65 & 89.05 & 4.39 \\
    \midrule
    VD-BERT
    & 59.96 & 65.44 & 51.63 & 82.23 & 90.68 &3.90 \\
    \midrule
    ICMU (Ours)
     &\bf 61.30 &\bf 66.82 &\bf 53.50 &\bf 83.05 &\bf 92.05 &\bf 3.59\\
    \bottomrule
    \end{tabular}}  
    \caption{Main comparisons on VisDial v1.0 test datasets (online). Our approach improves the strong baseline significantly. (t-test, p-value$ \textless$0.01)} 
    \label{tab:disc_test}
\end{table}

\subsection{Cross-Modal Training Objectives} \label{ssec:learning}
To make the model learn cross-modal understanding, we use two \textit{cross-modal} training losses---cross-modal masked token loss and cross-modal contrastive loss: 
\begin{equation}
    \mathcal{L} = \mathcal{L}_{CMTL} + \mathcal{L}_{CCL4},
\end{equation}
where $\mathcal{L}_{CMTL}$ is the cross-modal masked token loss and $\mathcal{L}_{CCL4}$ is a novel 4-way contrastive loss.

\subsubsection{Cross-modal Masked Token Loss}
At each iteration, we randomly mask each input token with probability $15\%$ and replace the masked one with a special token \texttt{[MASK]}. The model is then required to recover them based not only on the surrounding tokens $\mathbf{w}_{\backslash m}$ but also on the image $I$ by minimizing the negative log-likelihood: 
\begin{equation}\label{eq:MLM}
  \mathcal{L}_{CMTL} = -E_{(I, \mathbf{w})\sim D}\log P(w_m| \mathbf{w}_{\backslash m}, I),
\end{equation}
where $w_m$ refers to the masked token and $D$ denotes the training set.

\subsubsection{Cross-modal Contrastive Loss}
As shown in \figref{fig:framework}, to compute contrastive losses, for each input quartette $X = (I, H, Q, A)$, we construct three types of negative (unmatched) quartettes, where $I$ denotes the image, $H$ denotes the history, $Q$ denotes the question, $A$ denotes the answer. The first one is the polluted image $(I^*, H, Q, A)$, the second is the polluted question $(I, H, Q^*, A)$ and the final one is the polluted answer $(I, H, Q, A^*)$, where $*$ denotes the polluted input. Since the encoding of \texttt{[CLS]} can be viewed as a representation of the quartette $X = (I, H, Q, A)$, we apply a fully-connected (FC) layer on top of it as a 4-way classifier $f(\cdot)$ to predict whether the quartette is matched (c = 0), contains a polluted ${I^*} (c = 1)$, or contains a polluted ${Q^*} (c = 2)$ or contains a polluted ${A^*} (c = 3)$. The 4-way contrastive loss is defined as
\begin{equation}\label{eq:NSP}
  \mathcal{L}_{CCL4} = -E_{({I, H, Q, A};c)\sim D}  \log P(c|f({I, H, Q, A}),
\end{equation}
where the datasets ${I, H, Q, A} \in D$ contains $50\%$ matched quartettes, and the three negatives evenly  divide the remaining $50\%$ in the training set.

\subsection{Using VQA to Enhance Visual Dialog}
Although VQA is single-turn, VQA models and visual dialog models require similar cross-modal understanding capabilities. We use VQA to enhance visual dialogue. We exploit the training and val split of VQA v2.0 dataset, which contains the same images as VisDial v1.0 train split. As there is no caption for the image in VQA v2.0, we use VisDial v1.0 to construct a caption for each image in the VQA v2.0. Thus each input from VQA v2.0 can be defined as $(I, C, Q, A)$, where $I$ denotes the image, $C$ denotes the constructed caption, $Q$ denotes the question, $A$ denotes the answer. We let the history $H$ be null.  

% \vspace{-0.2em}
% \subsection{Training Setting} 
% For training in the discriminative setting, we transform the task of selecting an answer into a pointwise binary classification problem. 
% Specifically, we sample an answer $\hat{A}_t$ from the candidate pool and append it to the input sequence, and ask the CCL4 to distinguish whether the sampled answer is correct or pulled.
% We employ the \textit{bidirectional} self-attention mask to allow all the tokens to attend to each other by setting the mask matrix $\mathbf{M}$ in Eq.~(\ref{eq:mask}) to all $0$s.
% To avoid imbalanced class distribution, we keep the ratio of positive and negative instances to 1:1 in each epoch.
% To encourage the model to penalize more on negative instances, we randomly resample a negative example from the pool of $99$ negatives w.r.t. every positive one at different epochs.
% During inference, we rank the answer candidates according to the class score $c=3$ of their CCL4.

\begin{table}[t!]
    \centering
    \resizebox{0.98\columnwidth}!{
    \begin{tabular}{l|cccccc}
    \toprule
     Model & NDCG & MRR & R@1 & R@5 & R@10 & Mean \\
     \midrule
     MN & - & 60.29 & 46.14 & 77.68 & 87.57 & 4.84 \\
     HCIAE & - & 61.96 & 48.25 & 78.97 & 88.43 & 4.56 \\
     CoAtt & -  & 62.77 & 49.38 & 78.99 & 88.49 & 4.56 \\
     ReDAN & -  & 64.29 & 50.65 &  81.29 & 90.17 & 4.10 \\
     KBGN & 59.08 & 64.86 & 51.37 & 81.71 & 90.54 & 4.00  \\
     LG& 59.67 &  65.03  &  51.69  & 81.49 & 90.32 & 4.02 \\
     GoG& 63.15 & 62.68 & 49.46 & 78.77 & 87.87 & 4.81 \\
        \midrule
    VisDial-BERT  & 62.64  & 67.86  & 54.54  & 84.34 & 92.36 & 3.44 \\
     VD-BERT  & 63.22  & 67.44  & 54.02  & 83.96 & 92.33 & 3.53 \\
     \midrule
     ICMU (Ours) & \bf 64.30 & \bf 69.14 & \bf 56.80 &  \bf 85.09 & \bf 93.42 & \bf 3.37\\
    \bottomrule
    \end{tabular}}  
    \caption{Main comparisons on VisDial v1.0 val datasets. Our approach improves the strong baseline significantly. (t-test, p-value$ \textless$0.01)} 
    \label{tab:disc_val}
\end{table}

% \begin{table}[t!]
%     \centering
%     \resizebox{0.98\columnwidth}!{
%     \begin{tabular}{l|cccccc}
%     \toprule
%      Model & NDCG & MRR & R@1 & R@5 & R@10 & Mean \\
%      \midrule
%      MN~\cite{das2017visual}& - & 60.29 & 46.14 & 77.68 & 87.57 & 4.84 \\
%      HCIAE~\cite{lu2017best} & - & 61.96 & 48.25 & 78.97 & 88.43 & 4.56 \\
%      CoAtt~\cite{wu2018you} & -  & 62.77 & 49.38 & 78.99 & 88.49 & 4.56 \\
%      ReDAN~\cite{gan2019multi} & -  & 64.29 & 50.65 &  81.29 & 90.17 & 4.10 \\
%      KBGN~\cite{jiang2020kbgn} & 59.08 & 64.86 & 51.37 & 81.71 & 90.54 & 4.00  \\
%      LG~\cite{chen2021learning} & 59.67 &  65.03  &  51.69  & 81.49 & 90.32 & 4.02 \\
%      GoG~\cite{Chen2021GoGRG} & 63.15 & 62.68 & 49.46 & 78.77 & 87.87 & 4.81 \\
%         \midrule
%      VD-BERT~\cite{wang2020vd}  & 63.22  & 67.44  & 54.02  & 83.96 & 92.33 & 3.53 \\
%      \midrule
%      ICMU (Ours) & \bf 64.30 & \bf 69.14 & \bf 56.80 &  \bf 85.09 & \bf 93.42 & \bf 3.37\\
%     \bottomrule
%     \end{tabular}}  
%     \caption{Main comparisons on VisDial v1.0 val datasets. Our approach improves the strong baseline significantly. (t-test, p-value$ \textless$0.01)} 
%     \label{tab:disc_val}
% \end{table}

\begin{figure*}[t]
\centering
\scalebox{0.90}{
  \begin{overpic}[width=\textwidth]{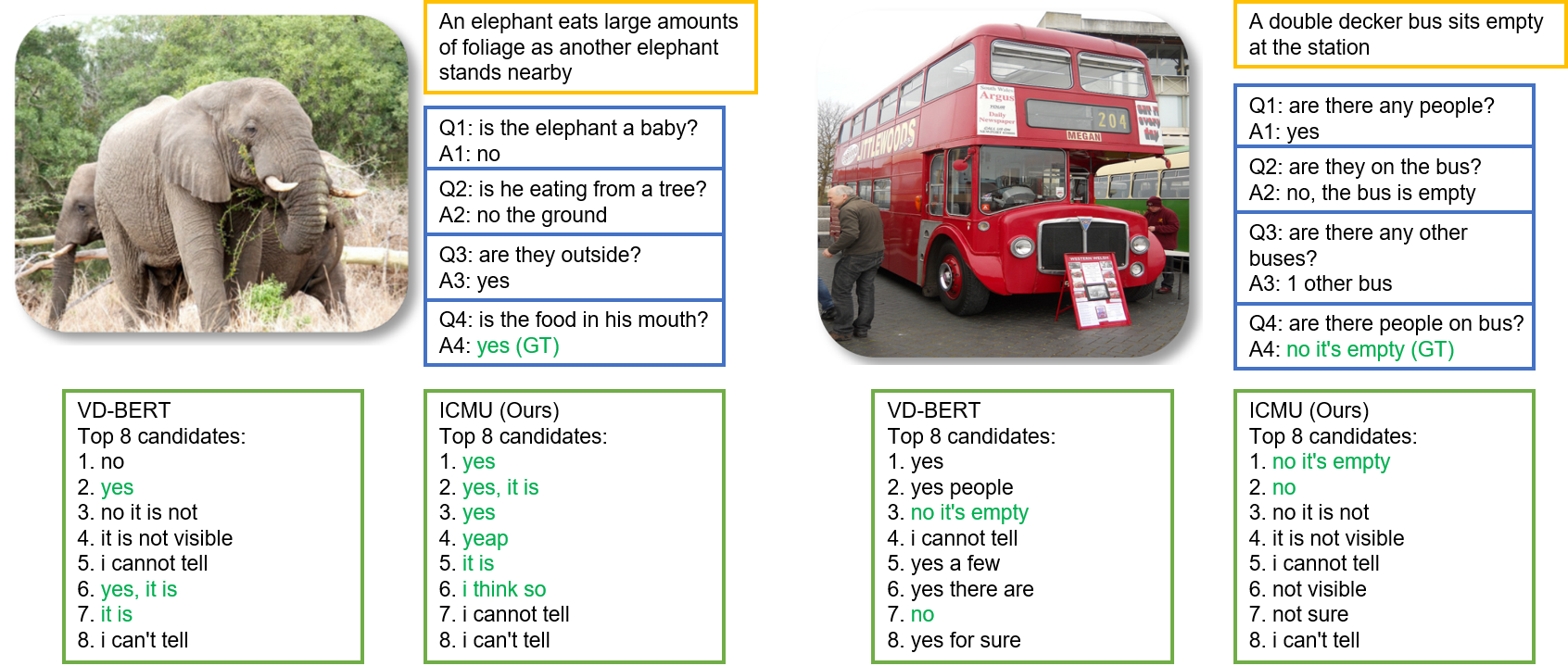}
  \end{overpic}
  }
  \caption{Case study.
  }\label{fig:case} 
\end{figure*}

\begin{table}[t!]
    \centering
    \resizebox{0.98\columnwidth}!{
    \begin{tabular}{l|cccccc}
    \toprule
     Model & NDCG & MRR & R@1 & R@5 & R@10 & Mean \\
     \midrule
     ICMU & \bf 64.30 & \bf 69.14 & \bf 56.80 & \bf 85.09 & \bf 93.42 & \bf 3.37\\
     \quad - VQA & 63.32 & 67.62 & 54.50 & 84.10 & 92.90 & 3.44 \\
     \quad - CL  & 63.34 & 67.90 & 54.82 & 84.35 & 92.43 & 3.52  \\
    %  \quad - BOTH & 57.06 & 63.32 & 48.80 & 81.10 & 89.60 & 4.14 \\
    \bottomrule
    \end{tabular}}  
    \caption{Ablation study on VisDial v1.0 val datasets. ``VQA'' denotes enhancing by utilizing VQA. ``CL'' denotes the 4-way contrastive learning.} 
    \label{tab:disc_val_}
\end{table}
\vspace{-0.2em}
\section{Experiments}
\subsection{Experiment Setup}
\subsubsection{Datasets and Implementation Details.}
We evaluate our model on the VisDial v$1.0$ datasets~\cite{DBLP:conf/cvpr/DasKGSYMPB17}.
% \footnote{Available at \url{https://visualdialog.org/data}}. 
Specifically, v$1.0$ contains a training set of 123287 images, a validation set of 2048 images and a testing set (hosted blindly in the task organizers' server) of 8,000 images. Each image is associated with one caption and 10 question-answer pairs. For each question, it is paired with a list of 100 answer candidates, one of which is regarded as the correct answer. VQA v2.0 contains the same 123287 images as VisDial v.10 but different question-answer pairs.

% For the v$1.0$ validation split and a part of v$1.0$ train split (2,000 images), extra dense annotations for the answer candidates are provided to make the evaluation more reasonable. 
% The dense annotation specifies a relevance score for each answer candidate based on the fact that some candidates with similar semantics to the ground truth answer can also be considered as correct or partially correct, e.g., ``brown and tan'' and ``brown'' in Figure~\ref{fig:framework}.

We use BERT\textsubscript{BASE} as the backbone, which consists of $12$ Transformer blocks, each with 12 attention heads and a hidden state dimensions of $768$.   
We use Adam~\cite{DBLP:journals/corr/KingmaB14} with an initial learning rate of $3e-5$ and a batch size of $80$ to train our model.
A linear learning rate decay schedule with a warmup of $0.1$ is employed.
We first train our model for $20$ epochs on a cluster of $4$ A$100$ GPUs with $40$G memory using CMTL and CCL4 losses (with equal coefficients).
Here we only utilize one previous dialog turn for training efficiency.
After that, we train for another $15$ epochs only using CCL4 losses. During inference, we rank the answer candidates according to the class score $c=0$ of the CCL4 loss.

% We use BERT\textsubscript{BASE} as the backbone, which consists of $12$ Transformer blocks, each with 12 attention heads and a hidden state dimensions of $768$.   
% We keep the max input sequence length (including $36$ visual objects) to $250$.
% We use Adam~\cite{DBLP:journals/corr/KingmaB14} with an initial learning rate of $3e-5$ and a batch size of $32$ to train our model.
% A linear learning rate decay schedule with a warmup of $0.1$ is employed.
% We first train VD-BERT for $30$ epochs on a cluster of $4$ V$100$ GPUs with $16$G memory using MLM and NSP losses (with equal coefficients).
% Here we only utilize one previous dialog turn for training efficiency.
% For instances where the appended answer candidate is incorrect, we do not conduct MLM on the answer sequence to reduce the noise introduced by the negative samples.
% After that, we  train for another $10$ epochs with full dialog history using either NSP in the discriminative setting or MLM on the answer sequence in the generative setting.
% For dense annotation fine-tuning in the discriminative setting, we train with the ListNet loss for $5$ epochs.
\vspace{-0.2em}
\subsubsection{Automatic Evaluation}
We use a retrieval setting to evaluate individual responses at each round of a dialog, following~\cite{das2017visual}. Specifically, at test time, apart from the image, ground truth dialog history and the question, a list of 100-candidate answers is also given. The model is evaluated on retrieval metrics: (1) Mean Rank of human response (Mean $\downarrow$), (2) Existence of the human response in $top-k$ ranked responses, i.e., R@$k$ $\uparrow$ (3) Mean Reciprocal Rank (MRR $\uparrow$) of the human response and (4) Normalized Discounted Cumulative Gain (NDCG $\uparrow$) for VisDial v1.0.

\vspace{-0.2em}
\subsection{Main Results}
\subsubsection{Baseline Methods}
We compare our method with the following baseline methods: (1) Attention-based models: HCIAE~\cite{lu2017best}, CoAtt~\cite{wu2018you}, ReDAN~\cite{gan2019multi}, LG~\cite{chen2021learning}. (2) The pretraining model: VD-BERT~\cite{wang2020vd} and VisDial-BERT~\cite{murahari2019large}. (4) Graph-based models: GNN-EM~\cite{zheng2019reasoning}, DualVD~\cite{jiang2020dualvd}, FGA~\cite{schwartz2019factor}, GoG~\cite{Chen2021GoGRG}, KBGN~\cite{jiang2020kbgn}.
% Please refer to Appendix \textcolor{red}{A.3} for more compared methods. 
% Since there are too many previous models, we compare some recent better models. We compare our method with the following baseline methods: (1) Attention-based models: HCIAE~\cite{lu2017best}, CoAtt~\cite{wu2018you}, Primary~\cite{guo2019image}, ReDAN~\cite{gan2019multi}, CorefNMN~\cite{Kottur2018VisualCR}, RvA~\cite{niu2019recursive}, DVAN~\cite{guo2019dual} and DMRM~\cite{chen2020dmrm}, DAM~\cite{jiang2020dam}. (3) The pretraining model: VD-BERT~\cite{wang2020vd} and VisDial-BERT~\cite{murahari2019large}. (4) Graph-based models: GNN~\cite{zheng2019reasoning}, DualVD~\cite{jiang2020dualvd}, FGA~\cite{schwartz2019factor},
% KBGN~\cite{jiang2020kbgn}. Note that VisDial-BERT\cite{murahari2019large} uses large extra 3.3 million image-caption pairs to learn cross-modal understanding. For fair comparison, we do not compare with VisDial-BERT.

\vspace{-0.2em}
\subsubsection{Results}
Performance on the benchmarks VisDial is shown in~\tabref{tab:disc_test} and~\tabref{tab:disc_val}. From the results on VisDial v1.0 test shown in \tabref{tab:disc_test}, we can observe that: 
(1) ICMU outperforms previous works on all metrics and obtains R@1 at 53.50\%, beating the previous method VD-BERT by $1.47\%$, which shows that ICMU can select the standard ground-truth more accurate.
(2) Comparing the performance of ICMU and model VD-BERT on NDCG, ICMU beats the pre-trained model VD-BERT by 1.34\%. This shows the superiority of our proposed method to understand cross-modal information at a fine-grained level. Note that NGCG is invariant to the order of options with identical relevance and to the order of options outside of the top K, where K is the number of answers marked as correct by at least one annotator.
(3) Our approach is not only more accurate (R@1, Mean), but also better than previous models on multi-modal semantic understanding (NDCG).

From the results on VisDial v1.0 val shown in \tabref{tab:disc_val}, we can get the same observations. From the ablation study on VisDial v1.0 val shown in \tabref{tab:disc_val_}, we can observe that: 
(1) Both cross-modal contrastive learning and enhancement by VQA bring satisfactory improvements.
(2) cross-modal contrastive learning and enhancement by VQA can get along with each other and further improve the performance of the model.

\vspace{-0.2em}
\subsubsection{Case Study}
As shown in~\figref{fig:case}, we provide two samples to analyze the cross-modal understanding of VD-BERT and ICMU. As shown in the left half of \figref{fig:case}, for Q4 ``\texttt{Does he have food in his mouth?}'', there are many reasonable answers to this question. VD-BERT ranks the opposite answer \texttt{``no''} first, and many reasonable answers ``\texttt{yes, it is, it is}'' are ranked lower. As shown in the right half of \figref{fig:case}, for Q4 ``\texttt{are there people on bus?}'', ICMU outperforms the VD-BERT. This shows that ICMU learns better cross-modal understanding than VD-BERT due to CCL4 and enhancing by VQA.

\vspace{-0.2em}
\section{Conclusion}
In this paper, we propose a novel approach to improve the cross-modal understanding for visual dialog, named ICMU. ICMU enhances the cross-modal understanding in visual dialog by distinguishing different pulled inputs based on 4-way contrastive learning.  In addition, ICMU exploits the single-turn visual question answering to enhance the visual dialog model's cross-modal understanding. Experiments show that the proposed approach improves the visual dialog model's cross-modal understanding and brings satisfactory gain to the VisDial dataset.

\vspace{-0.2em}
\section{Acknowledgement}
This work was supported by the National Key R\&D Program of China under Grant No.2018YFB1005104 and the Key Research Program of the Chinese Academy of Sciences under Grant No.ZDBS-SSW-JSC006 and Strategic Priority Research Program of the Chinese Academy of Sciences under Grant No.XDA27030300.

\newpage
\small
\bibliographystyle{IEEEbib}
\bibliography{strings,refs}

\end{document}